\pdfoutput=1

\documentclass[11pt]{article}
\usepackage{acl}
\usepackage{times}
\usepackage{latexsym}

\usepackage[T1]{fontenc}
\usepackage[utf8]{inputenc}
\usepackage{microtype}
\usepackage{inconsolata}

\usepackage{amsthm}

\usepackage{amsmath}
\usepackage{graphicx}
\usepackage{algorithm}
\usepackage[noend]{algpseudocode}

\usepackage{enumitem}

\usepackage{multirow}
\usepackage{adjustbox}
\usepackage{bbm}
\usepackage{wrapfig,lipsum}
\usepackage{xspace}
\usepackage{booktabs,cellspace}  
\usepackage[normalem]{ulem}
\usepackage{makecell}
\usepackage{amssymb}
\usepackage{pifont}
\usepackage{todonotes}
\newcommand{\xmark}{\ding{55}}%

\usepackage{microtype}

\usepackage{cleveref}

\newcommand{\stitle}[1]{\vspace{1ex} \noindent{\bf #1.}}

\newcommand{\modelname}{\textsc{DeepEdit}\xspace}
\newcommand{\cona}{\textsc{Conciseness}\xspace}
\newcommand{\conb}{\textsc{Coherence}\xspace}
\newcommand{\conc}{\textsc{Receptiveness}\xspace}
\newcommand{\cond}{\textsc{Pertinence}\xspace}
\newcommand{\dataa}{\textsc{MQuAKE}\xspace}
\newcommand{\datab}{\textsc{MQuAKE-3k}\xspace}
\newcommand{\datac}{\textsc{MQuAKE-2002}\xspace}
\newcommand{\datad}{\textsc{MQuAKE-hard}\xspace}
\newcommand{\datae}{\textsc{CounterFact}\xspace}
\newcommand{\llamaa}{\textsc{LLaMA2-7B-Chat}\xspace}
\newcommand{\gpta}{\textsc{GPT-3.5-Turbo-Instruct}\xspace}
\newcommand{\gptb}{\textsc{Text-Davinci-003}\xspace}

\usepackage{etoolbox}
\usepackage[tikz]{bclogo}

\usepackage{subcaption}
\usepackage{adjustbox}

\definecolor{msftBlue}{RGB}{0,164,239}
\definecolor{msftGreen}{RGB}{127,186,0}
\definecolor{msftYello}{RGB}{255,185,0}
\definecolor{msftBlack}{RGB}{0,0,0}

\usepackage{tcolorbox} 
\tcbuselibrary{skins} 
\usepackage[T1]{fontenc}

\tcbset{
    userstyle/.style={
        enhanced,
        colback=white,
        colframe=black,
        colbacktitle=gray!20,
        coltitle=black,
        rounded corners,
        sharp corners=north,
        boxrule=0.5pt,
        drop shadow=black!50!white,
        attach boxed title to top left={
            xshift=-2mm,
            yshift=-2mm
        },
        boxed title style={
            rounded corners,
            size=small,
            colback=gray!20
        }
    },
    replystyleg/.style={
        enhanced,
        colback=green!15,
        colframe=black,
        colbacktitle=green!30,
        coltitle=black,
        boxrule=0.5pt,
        drop shadow=black!50!white,
        rounded corners,
        sharp corners=north,
        attach boxed title to top right={
            xshift=-2mm,
            yshift=-2mm
        },
        boxed title style={
            rounded corners,
            size=small,
            colback=green!40
        }
    },
    replystyler/.style={
        enhanced,
        colback=red!15,
        colframe=black,
        colbacktitle=red!40,
        coltitle=black,
        boxrule=0.5pt,
        drop shadow=black!50!white,
        rounded corners,
        sharp corners=north,
        attach boxed title to top right={
            xshift=-2mm,
            yshift=-2mm
        },
        boxed title style={
            rounded corners,
            size=small,
            colback=red!40
        }
    }
}

\newtcolorbox{userquery}[1][]{
    userstyle,
    title=Prompt,
    #1
}

\title{DeepEdit: Knowledge Editing as Decoding with Constraints}

\author{
Yiwei Wang$^\dagger$ \ \ \ \ Muhao Chen$^\ddagger$ \ \ \ \ Nanyun Peng$^\dagger$ \ \ \ \ Kai-Wei Chang$^\dagger$ \\ 
$^\dagger$ University of California, Los Angeles \quad $^\ddagger$ University of California, Davis \\
\texttt{wangyw.evan@gmail.com}
\\
\href{https://wangywust.github.io/deepedit.io/}{\textcolor{magenta}{\texttt{https://wangywust.github.io/deepedit.io/}}}
\url{}
}

\date{}

\begin{document}
\maketitle
\begin{abstract}
Incorporating new knowledge in multi-step reasoning has become a major problem in the research of knowledge editing (KE).
LLMs often struggle to generate coherent and precise reasoning chains when incorporating new in-context knowledge, as they tend to deviate from relevant information and lose logical coherence during reasoning.
To address this issue, it is crucial to apply direct controls over LLM outputs to ensure both logical coherence and effective integration of new knowledge into their reasoning processes.
To this end, we propose \textbf{\modelname} (\textit{Depth-first Search-based Constrained Decoding for Knowledge Editing}), a KE method designed to enhance LLMs' reasoning capabilities with new knowledge.
\modelname employs the decoding constraints that we carefully crafted to explicitly improve logical coherence and knowledge integration during reasoning.
Additionally, we introduce a semantic-based importance estimator that ranks the relevance of reasoning steps to optimize the informativeness of output reasoning relative to the input question.
Beyond \modelname, we propose two new KE benchmarks, \textbf{\datac} and \textbf{\datad}, which offer more precise and challenging assessments of KE methodologies.
Qualitative analysis shows that \modelname enables LLMs to generate succinct and coherent reasoning chains incorporating new knowledge.
Quantitatively, it yields significant improvements on multiple KE benchmarks.
\end{abstract}

\section{Introduction}\label{sec:int}
\begin{figure*}[!tb]
	\centering
	\includegraphics[width=1\linewidth]{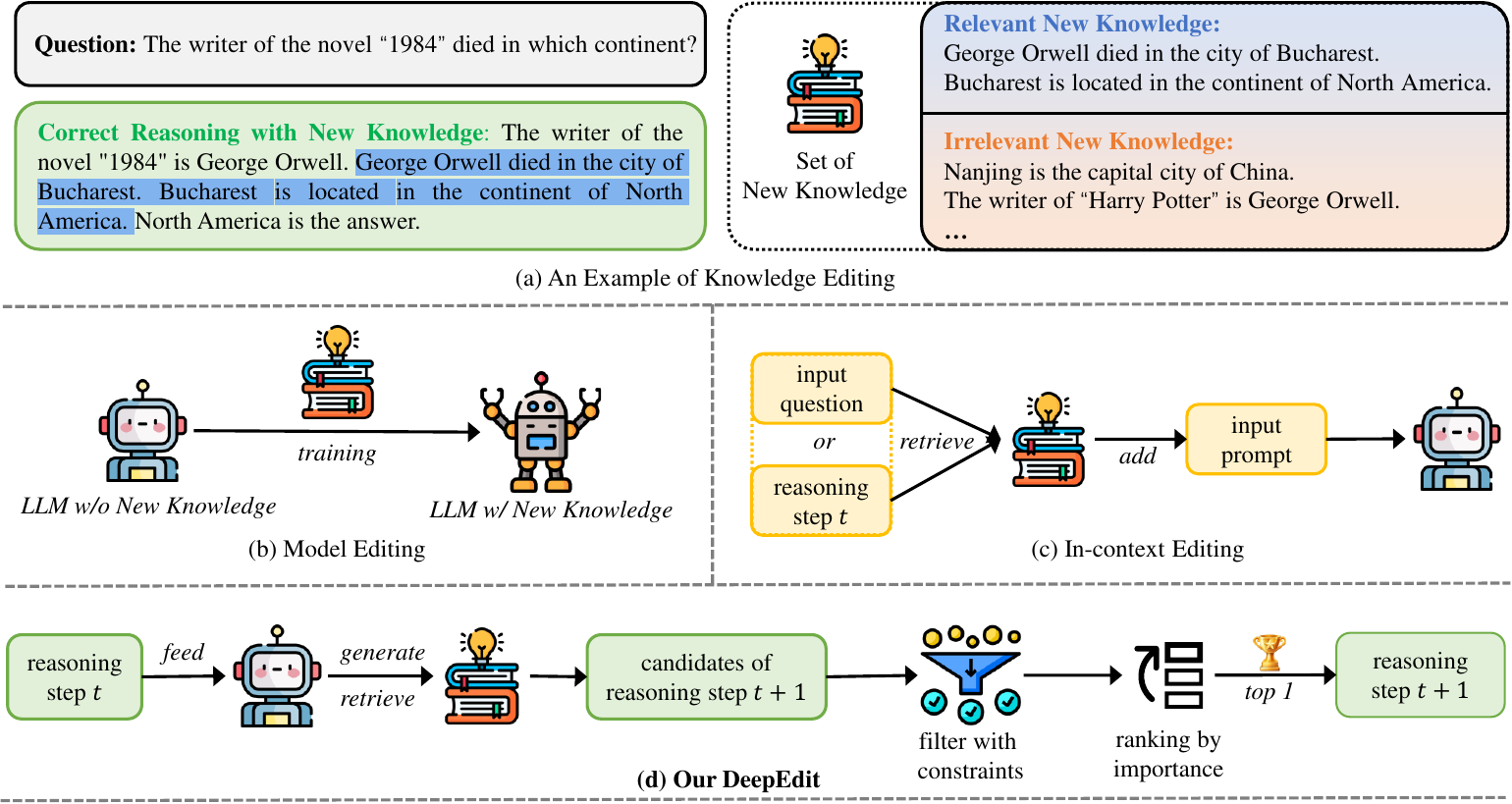}
	\caption{
    (a) An instance in KE evaluation benchmarks.
    (b,c) Prior KE methods that retrain LLMs with new knowledge (model editing) and adds new knowledge in the input prompt (in-context editing). 
    (c) Our \modelname directly controls LLMs' outputs to select the valid and most important (new) knowledge for reasoning.
    \vspace{-4mm}
\label{fig:all}}
\end{figure*}

The research of knowledge editing (KE) aims to enable large language models (LLMs) to be such question-answering systems that the outputs are receptive to the new knowledge that is different from the LLMs' parametric knowledge.
Multi-hop question answering with new knowledge is currently the major task to evaluate KE~\cite{zhong2023mquake}, which requires models to answer questions corresponding to a chain of facts including new knowledge (see \Cref{fig:all}(a)).

\begin{figure}[!t]
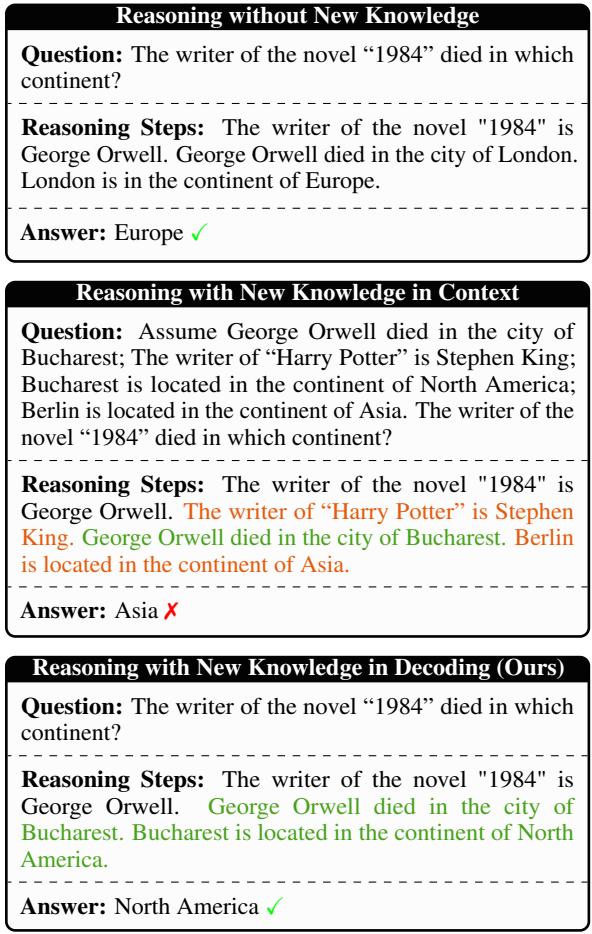

    \centering
\begin{tcolorbox}[fonttitle = \small\bfseries, title=Reasoning without New Knowledge,colframe=gray!2!black,colback=gray!2!white,boxrule=1pt,boxsep=0pt,left=5pt,right=5pt,fontupper=\footnotesize, halign title = flush center]
\textbf{Question:} The writer of the novel “1984” died in which continent?
\tcbline
\textbf{\color[RGB]{0,0,0}{Reasoning Steps}:} The writer of the novel "1984" is George Orwell. George Orwell died in the city of London. London is in the continent of Europe.
\tcbline
\textbf{\color[RGB]{0,0,0}{Answer}:} Europe {\color[RGB]{0,255,0}{\checkmark}}
\end{tcolorbox}
\begin{tcolorbox}[fonttitle = \small\bfseries, title=Reasoning with New Knowledge in Context,colframe=gray!2!black,colback=gray!2!white,boxrule=1pt,boxsep=0pt,left=5pt,right=5pt,fontupper=\footnotesize, halign title = flush center]
\textbf{Question:} Assume George Orwell died in the city of Bucharest; The writer of ``Harry Potter'' is Stephen King; Bucharest is located in the continent of North America; Berlin is located in the continent of Asia. The writer of the novel “1984” died in which continent?
\tcbline
\textbf{\color[RGB]{0,0,0}{Reasoning Steps}:} The writer of the novel "1984" is George Orwell. {\color[RGB]{229,91,11}{The writer of ``Harry Potter'' is Stephen King.}} {\color[RGB]{71,162,26}{George Orwell died in the city of Bucharest.}} {\color[RGB]{229,91,11}{Berlin is located in the continent of Asia.}}
\tcbline
\textbf{\color[RGB]{0,0,0}{Answer}:} Asia {\color[RGB]{255,0,0}{\xmark}}
\end{tcolorbox}
\begin{tcolorbox}[fonttitle = \small\bfseries, title=Reasoning with New Knowledge in Decoding (Ours),colframe=gray!2!black,colback=gray!2!white,boxrule=1pt,boxsep=0pt,left=5pt,right=5pt,fontupper=\footnotesize, halign title = flush center]
\textbf{Question:} The writer of the novel “1984” died in which continent?
\tcbline
\textbf{\color[RGB]{0,0,0}{Reasoning Steps}:} The writer of the novel "1984" is George Orwell. {\color[RGB]{71,162,26}{George Orwell died in the city of Bucharest. Bucharest is located in the continent of North America.}}
\tcbline
\textbf{\color[RGB]{0,0,0}{Answer}:} North America {\color[RGB]{0,255,0}{\checkmark}}
\end{tcolorbox}
    \caption{(\textit{Upper}) It is easy for LLMs to generate coherent reasoning chains without new knowledge.
    (\textit{Middle}) Given new knowledge in context, LLMs can hardly generate coherent reasoning chains due to hallucinations on new knowledge.
    As a result, both {\color[RGB]{71,162,26}{relevant}} and {\color[RGB]{229,91,11}{irrelevant}} knowledge participate in the reasoning.
    (\textit{Lower}) 
    We do not put new knowledge in context but let them directly contribute to the decoding, which leads to coherent and precise reasoning chains.
    }
    \label{fig:illu}
\end{figure}

Existing literature 
has mainly focused on performing KE through model editing \cite{mitchell2021fast,meng2022mass} and in-context editing \cite{zhong2023mquake,cohen2023evaluating,zheng2023can}. 
The former uses retraining to inject new knowledge into LLMs' weights while the latter adds new knowledge to the input prompt to fulfill KE (see \Cref{fig:all}(b)).  
Both of them lack direct control of LLMs' outputs, which is essential to enhance LLMs' multi-step reasoning incorporating new knowledge (see \Cref{fig:illu}). 

\begin{figure*}[!tb]
	\centering
	\includegraphics[width=1\linewidth]{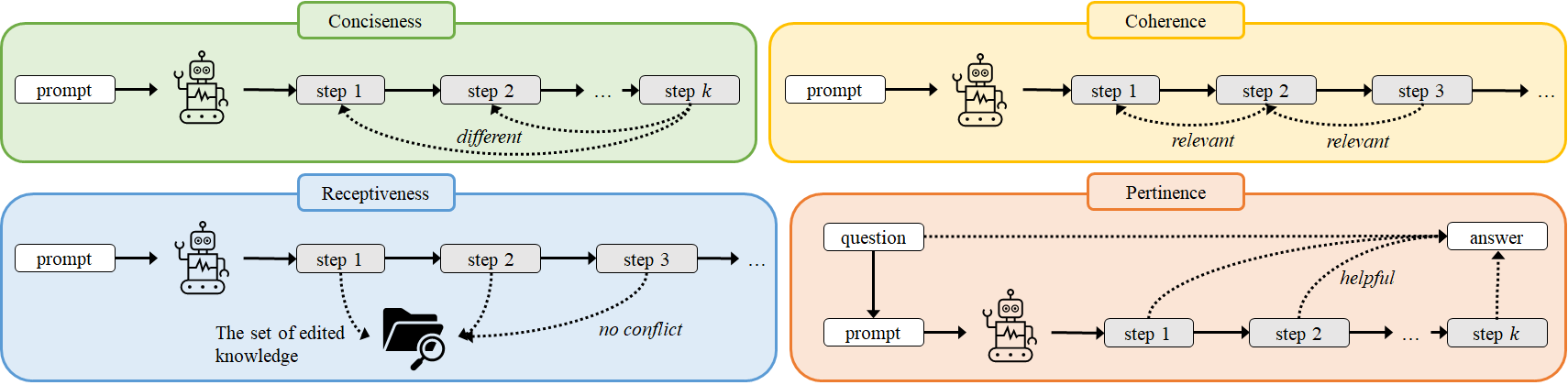}
	\caption{
    Our \modelname controls LLMs' reasoning to follow the constraints of \cona, \conb, \conc, and \cond so as to soundly incorporate new knowledge into LLMs' question answering.
\label{fig:con}}
\vspace{-2mm}
\end{figure*}

A satisfactory multi-step reasoning chain with new knowledge should consist of coherent steps and place new knowledge at appropriate positions (see Fig. \Cref{fig:all}(a).
In this work, we explore to propose a new constrained decoding algorithm that enhances LLMs' to produce such reasoning chains.
First, we design the decoding constraints that facilitate LLMs to soundly incorporate new knowledge in multi-step reasoning.
Specifically, we propose the following constraints: \cona requires every reasoning step to be identical; \conb requires the adjacent reasoning steps to be coherent; \conc requires new knowledge to replace the conflicted parametric knowledge; \cond requires every reasoning step to be relevant to the target question (see \Cref{fig:con}).

We incorporate the above constraints into a new KE method: \textbf{\modelname} (\textit{Depth-first Search-based Constrained Decoding for Knowledge Editing}).
When producing every reasoning step, \modelname searches through the parametric and new knowledge to find the knowledge items that satisfy our constraints, termed as valid step candidates.
Among the valid candidates, \modelname prioritizes more important ones as the new reasoning step for further reasoning.
We comprehensively evaluate the importance of different candidates based on their semantic relationships and the prior that the new knowledge is more informative to LLMs than the parametric knowledge.

Classical depth-first search has a backtracking mechanism that enables us to traverse all the reasoning chains of valid reasoning steps \cite{tarjan1972depth}. 
Such backtracking is time-consuming and unnecessary after an answer is found.
Therefore, we design an early-stopping mechanism to stop the depth-first search when an answer is found by LLMs. 
Such an early stopping significantly improves the reasoning efficiency by reducing redundant iterations, as validated by the qualitative analysis (see \Cref{sec:3_5}) and empirical results (see \Cref{sec:eff_exp}).

\modelname is flexibly applicable to any black-box LLM that only requires access to output texts.
Besides, \modelname improves both the receptiveness of new knowledge and the coherence of multi-step reasoning by directly controlling the LLMs' reasoning with decoding constraints.
Unlike previous KE methods that edit either the model parameters \cite{zhu2020modifying} or the input prompts \cite{cohen2023evaluating}, we directly control the output of LLMs to incorporate the new knowledge, which opens up a new direction for KE in LLMs.

To provide precise and challenging assessments of KE methods, we build two new benchmarks, \textbf{\datac}, and \textbf{\datad}, which resolve the knowledge-conflicting annotation mistakes in the popular KE benchmark \datab \cite{zhong2023mquake}.
We evaluate \modelname on \datab, \datac, and \datad \cite{zhong2023mquake,meng2022locating}. 
Qualitatively, \modelname enhances LLMs to produce succinct reasoning with new knowledge, as shown in \Cref{fig:case}.
Quantitatively, \modelname improves the KE for popular LLMs by a significant margin \cite{chatgpt,touvron2023llama}.

\section{Related Work}


\paragraph{Knowledge Editing}
Previous studies have explored multiple methods for editing the knowledge of LLMs by introducing new knowledge into static model artifacts \cite{zhu2020modifying,sotoudeh2019correcting,dai2021knowledge,hase2021language,zhou2023context,dong2022calibrating,huang2023transformer}. 
Some of these approaches involve identifying and modifying model weights associated with specific concepts \cite{meng2022locating,meng2022mass,dai2021knowledge}, as well as rapid adaptation facilitated by a compact auxiliary editing network \cite{mitchell2022memory,de2021editing,mitchell2021fast}. 
Given the fast-growing parameter sizes of large language models (LLMs) \cite{zhao2023survey}, frequently updating LLMs with new knowledge through retraining is more and more expensive.
Hence, it is vital to 
effectively edit the LLMs' knowledge without retraining.
Recent work observes the superior KE performance of in-context editing. 
MeLLo \cite{zhong2023mquake} designs a single prompt to alternately conduct the text generation and new knowledge analysis using the LLM generator.
\citet{cohen2023evaluating} propose to append the new knowledge the head of the input prompt.
In-context editing lacks direct control of the LLMs' outputs and how the new knowledge influences the outputs is not traceable.

Different from the prior work, we propose a new KE method \modelname, which effectively augments LLMs on KE with constrained decoding.
Our \modelname outperforms the prior KE methods by editing the LLMs' outputs directly, which paves the way to a new direction of LLMs' KE.


\paragraph{Constrained Decoding}
Constrained decoding \cite{qin2020back,qin2022cold,lu2020neurologic,lu2021neurologic,kumar2022gradient,liu2023bolt,geng2023flexible} is a classical natural language processing topic, which is mainly applied to control the lexicon, sentiment or style of the generated text. 
FUDGE \cite{yang2021fudge} employs weighted decoding (WD) by directly adjusting the probabilities of the vocabulary distribution with an auxiliary classifier probability.
NADO \cite{meng2022controllable} suggests sampling from a vocabulary distribution similar to an approximation of the sequence-level oracle.
GeDi \cite{krause2020gedi} and
DExperts \cite{liu2021dexperts} also adopt the weighted decoding approach through the training of generative classifiers.
Constrained decoding methods generally require access to the LLMs' token-wise distribution.
Different from them, our \modelname avoids the requirements of distribution access by decoding at the reasoning step level.
We extend the applications of constrained decoding to KE.
We design the semantic constraints to soundly incorporate new knowledge in LLMs' reasoning, and verify the complex semantic constraints with LLMs. 


		
		

\section{Methodology}
Multi-hop question answering with new knowledge assumes a database that stores all new facts, which LLMs should adhere to, and posits that only a few facts can influence the ground-truth answer to each question.
In this section, we will introduce how we integrate the decoding constraints into the design of a novel KE method, \modelname (Depth-first Search based Decoding for Knowledge Editing), to enhance the KE of black-box LLMs.

\subsection{Constraint Designing for Knowledge Editing} \label{sec:3_2}

On an LLM's multi-step reasoning chain to answer a multi-hop question involving new knowledge, we find the following constraints on reasoning steps can help LLMs to give the correct answer:
\begin{enumerate}[leftmargin=1em]
    \item \textbf{\cona}: Every step is unique.
    \item \textbf{\conb}: Adjacent steps are coherent.
    \item \textbf{\conc}: No conflicts exist between reasoning steps and new knowledge.
    \item \textbf{\cond}: Every reasoning step should be relevant to the target question.
\end{enumerate}

Among the above four constraints, \cona removes loops in the reasoning chain;
\conb guarantees the coherence of reasoning;
\conc ensures the LLMs' receptiveness to new knowledge;
\cond guides LLMs to answer the target question.
We visualize different constraints in \Cref{fig:con}.
The above constraints are easy to follow for LLMs when there is no new knowledge.
However, the utilization of in-context new knowledge is influenced by the LLMs' hallucinations, which can easily break the aforementioned constraints.
In this work, we apply the above constraints to incorporate the new knowledge into LLMs' reasoning.

\subsection{Constraint Verification on Every Step}
We term the reasoning steps that satisfy all the constraints as valid steps.
Because an invalid reasoning step would mislead LLMs into generating incorrect subsequent steps, we verify the constraints at the decoding of each reasoning step.

At every iteration, we take both the parametric and new knowledge as the step candidates.
First, we let LLMs generate one reasoning step with temperature as $0$, which represents the parametric knowledge that the LLM believes to be most useful to answer the target question.
Second, we retrieve $N$ new knowledge closest to the generated reasoning step.
The distance between knowledge and the generated step is measured by the semantic distance of sentence-level embeddings given by a pre-trained BERT model \cite{devlin-etal-2019-bert}.
The closer the context of the new knowledge is to the generated step, the more likely it is that the new knowledge will be helpful in answering the target question.
We take these $N + 1$ knowledge as the reasoning step candidates.

We verify the constraints on the aforementioned step candidates to identify the valid ones.
In principle, every constraint's verification can be seen as a binary classification problem on a pair of sentences.
For example, \conb's verification can be completed by assessing whether a pair of reasoning steps is relevant to each other or not.
Therefore, one option to do constraint verification is to train a binary classifier for every constraint.
However, such a solution is not data or time-efficient.
In this work, we develop four specialized in-context learning-based LLM agents to act as the verifiers.
Every verifier's prompt includes $D$ positive and $N$ negative demonstrations randomly sampled from \dataa \cite{zhong2023mquake}.
In our experiments, we set $D = 2$ and $N = 2$ by default.

\subsection{Efficient Reasoning with Depth-first Search} \label{sec:3_5}
At every iteration, there may exist more than one valid step candidates.
We can consider each valid step candidate simultaneously as the new reasoning step and make further reasoning based on each of them separately.
This design can be seen as a breadth-first search of a valid multi-hop reasoning chain.
It is undoubtedly time-consuming because the time complexity grows exponentially with the number of reasoning steps.
This raises a natural question:

\textit{Can we increase the depth of reasoning chains more efficiently than the breadth-first search?}

\begin{figure}[!tb]
	\centering
	\includegraphics[width=1\linewidth]{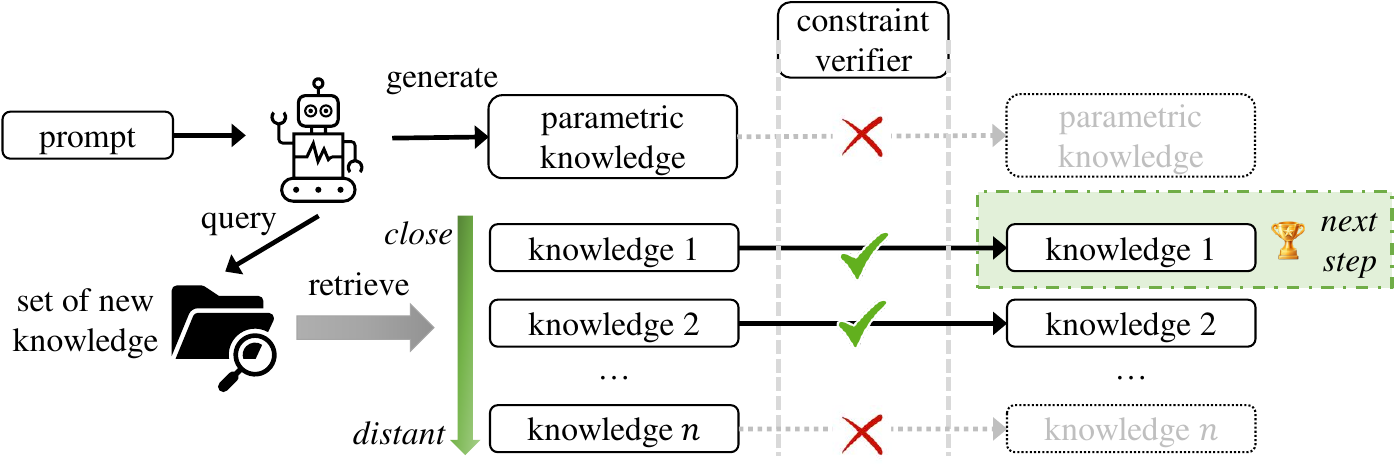}
	\caption{At every iteration, we verify the decoding constraints over parametric and new knowledge to find the valid step candidates. 
    Then we take the most important candidate as the next reasoning step to efficiently increase the reasoning depth.
\label{fig:global}}
\vspace{0mm}
\end{figure}

Our answer is yes.
We propose prioritizing more important step candidates as the reasoning step for further reasoning.
Actually, different valid step candidates are not equally important.
First, the edited facts should be more important than the generated step, since the former generally contains new knowledge unknown to LLMs.
Secondly, the edited facts that are closer to the generated step are more important. 
They are more helpful in answering the target question due to their higher relevance to the context generated by the LLM's reasoning.
Based on the above prior, in every iteration, we select the most important knowledge as the next reasoning step and do further reasoning based on it, as shown in \Cref{fig:global}.

\begin{figure*}[!tb]
	\centering
	\includegraphics[width=1\linewidth]{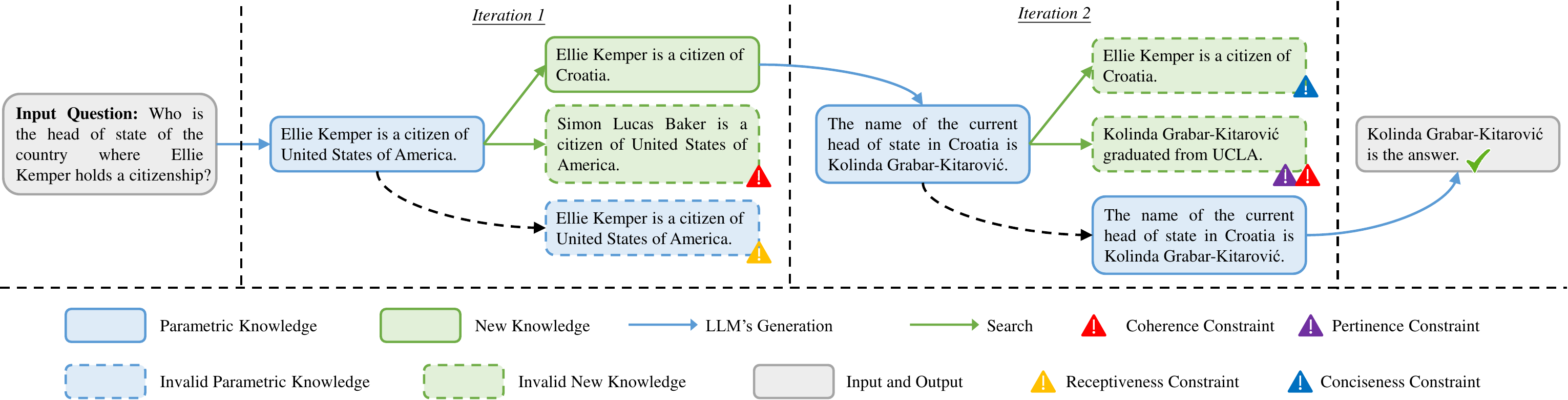}
	\caption{
    The iteration-wise visualization of our \modelname on an instance in \datab. 
    In every iteration, the retrieved new knowledge (highlighted in green) is ranked from top to bottom as step candidates based on their semantic distances to the generated step, in ascending order.
    \label{fig:iteration-wise}}
\vspace{0mm}
\end{figure*}

We name our method as \modelname: depth-first search-based constrained decoding for knowledge editing.
If we follow the classical depth-first search algorithm \cite{tarjan1972depth}, we should include a backtracking mechanism to traverse all the reasoning chains of valid steps.
However, such backtracking is undoubtedly time-consuming and unnecessary after an answer is found.
Therefore, we design an early-stopping mechanism to stop the search when an answer is found by LLMs.
We visualize \modelname in \Cref{fig:all} the example iteration-wise outputs in \Cref{fig:iteration-wise}.

Assume the number of reasoning hops as $h$, and the number of valid step candidates per iteration as $c$.
The time complexity of breadth-first search is $\mathcal{O}(c^h)$.
In contrast, thanks to our depth-first search design with the early-stopping mechanism, our model has a significantly lower time complexity of $\mathcal{O}(h)$ in the best case.
Empirical results validate \modelname's superior efficiency compared with the breadth-first search (see \Cref{sec:eff_exp}).

The strategy of depth-first search has been utilized by some research on LLMs, such as tree-of-thoughts \cite{yao2023tree}. 
Different from them, our \modelname's depth-first search is designed to efficiently find the valid multi-hop reasoning chain incorporating new knowledge.
Our \modelname is free to be applied to all black-box LLMs: it bypasses the requirement of accessing the token-wise distributions because our decoding strategy is defined at the reasoning step level.


\section{New Benchmarks for More Precise and Challenging Evaluation of KE}\label{sec:4}

We provide two new benchmarks for the evaluation of KE methods: \datac and \datad.
Both are built based on \dataa \cite{zhong2023mquake}, a recent KE dataset that is designed to evaluate the KE methods on helping LLMs to answer multi-hop questions given new knowledge.
Every instance in \dataa includes a multi-hop question and several edited facts, each of which can alter the ground-truth answer to the question.

\subsection{Annotation Mistakes in One-Third Instances of \dataa}

\begin{figure}[!tb]
	\centering
	\includegraphics[width=1\linewidth]{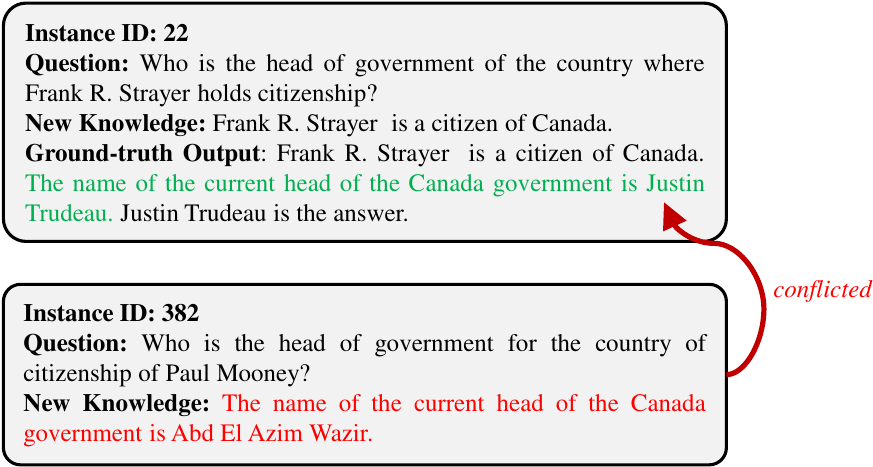}
	\caption{
    An example of knowledge conflicts across instances in \datab. 
    New knowledge (in red) in instance 382 is conflicted with the ground-truth reasoning step (in green) of instance 22.
    \vspace{-5mm}
\label{fig:conflict}}
\end{figure}

\citet{zhong2023mquake} suggests using a randomly sampled 3,000 instance subset of MQuAKE to do the evaluation, which reduces the experimental costs, known as \datab.
There are two issues of using \datab to evaluate KE.
The first issue is that the new knowledge from different instances \datab can cause conflicts and mistakes in the ground-truth answers.
In other words, the ground-truth answer from instance $A$ can be altered by the new knowledge from another instance $B$. 
We show an example of knowledge conflicts in \datab in \Cref{fig:conflict}.
These knowledge conflicts will make the ground-truth answers incorrect given the conflicted new knowledge because the inference on every instance could retrieve the new knowledge from all instances.
We observe that 998 instances' ground-truth answers are broken by the new knowledge conflicts across instances.

\subsection{New Benchmark \datac for More Precise Evaluation}
To address the issue of annotation mistakes in \datab, we provide a new KE benchmark based on \dataa, which does not have any knowledge conflict across instances.
This benchmark includes 2,002 instances, so we term it as \datac.
We filter out the instances of which the ground-truth answers are broken by the new knowledge from other instances to produce \datac.
Compared with \datab, our \datac provides a more precise evaluation for KE methods, since it removes the annotation mistakes due to knowledge conflicts across instances.
The data statistics of \datac is provided in \Cref{tab:data}.

\subsection{New Benchmark \datad for Challenging Evaluation}

The second issue of \datab is that more than 60\% instances in it only contain at most two edited facts that influence the answers, which are not challenging enough to evaluate the KE methods on handling multiple edited facts that can alter the ground-truth answers.
We construct a more challenging subset of \dataa by selecting the instances that contain the highest number of edited facts per instance excluding \datab.
We term this challenging set as \datad, which includes 429 instances and every instance contains four edited facts that can change the ground-truth answer. 
The data statistics of \datad is also provided in \Cref{tab:data}.

\begin{table*}[t]
	\centering
	\begin{adjustbox}{width=\linewidth}
		\begin{tabular}{@{}lcccc@{}}
			\toprule
			\textbf{Benchmark}
                & \# Instances
                & \# Hops per Instance
                & \# Edited Facts per Instance
                & \# Conflicted Instances
   \\ 
			\midrule
			\midrule
            \datab \cite{zhong2023mquake} & 3,000 & 3.0 & 2.0 & 998\\
	      \datac (Ours) & 2,002 & 2.7 & 2.2 & 0 \\
	      \datad (Ours) & 429 & 4.0 & 4.0 & 0 \\
			\bottomrule
		\end{tabular}
	\end{adjustbox}
	\caption{Data statistics of different benchmarks.
    \# Conflicted Instances represent the number of instances of which the ground-truth labels are affected by the new knowledge from other instances.
    An example is shown in \Cref{fig:conflict}.
 \vspace{0mm}
    \label{tab:data}}
\end{table*}

\section{Experiments}
In this section, we 
evaluate the KE performance of our \modelname method 
when applied to LLMs, and compare against strong KE baselines \cite{zhong2023mquake}.
Our experimental settings closely follow those of the previous work \cite{zhong2023mquake} to ensure a fair comparison.

\subsection{Experimental Settings}
We take the following KE benchmarks \datab \cite{zhong2023mquake}, 
\datac, and \datad for evaluation.
We follow the existing work \cite{zhong2023mquake} to use accuracy as the main evaluation metric. 
For every instance, we test the accuracy of the first question in the instance, since different questions in the same instance are semantically equivalent and correspond to the same ground-truth answer.
We follow \cite{zhong2023mquake} to set the KE batch size as 100 and full batch for evaluation.
The KE batch size means the number of instances that provide new knowledge for retrieval when answering each question.

\stitle{Compared methods} 
We take the strong KE methods into comparison: MEND \cite{mitchell2021fast}, MEMIT \cite{meng2022mass}, IKE \cite{zheng2023can}, ICE \cite{cohen2023evaluating}, and MeLLo \cite{zhong2023mquake}.
MEND produces weight updates by transforming the raw fine-tuning gradients given an edited fact.
MEMIT updates feedforward networks in a range of layers to encode the edited facts.
IKE and ICE conduct KE by adding new knowledge to the input prompts.
MeLLo is the state-of-the-art KE method that prompts the LLMs to generate subquestions and append new knowledge to the prompts.
The former two KE methods are only applicable to white-box LLMs, while the latter three can fit the block-box LLMs.

\begin{table*}[t]
	\centering
	\begin{adjustbox}{width=\linewidth}
        \setlength{\tabcolsep}{3pt}
		\begin{tabular}{@{}lcccccc@{}}
\toprule
\multirow{2}{*}{\textbf{Method}}
& \multicolumn{2}{c}{\textbf{\datab}}
&\multicolumn{2}{c}{\textbf{\datad}}
&\multicolumn{2}{c}{\textbf{\datac}}
\\
& 100 edited
& all edited
& 100 edited
& all edited
& 100 edited
& all edited
\\
\midrule
\midrule
\llamaa w/ MEND \cite{mitchell2021fast} & 7.2 & 3.9 & 0.5 & 0.2 & 7.7 & 4.1\\
\llamaa w/ MEMIT \cite{meng2022mass} & 7.9 & 4.2 & 0.9 & 0.5 & 8.3 & 4.8\\
\llamaa w/ IKE \cite{zheng2023can} & 12.5 & 6.2 & 1.2 & 0.5 & 12.9 & 6.5\\
\llamaa w/ ICE \cite{cohen2023evaluating} & 12.9 & 6.3 & 1.4 & 0.7 & 13.2 & 6.5\\
\llamaa w/ MeLLo \cite{zhong2023mquake} & 16.3 & 10.8 & 1.9 & 1.6 & 17.1 & 11.8\\
\llamaa w/ \modelname (Ours) & \textbf{27.6} & \textbf{11.2} & \textbf{14.5} & \textbf{7.0} & \textbf{29.5} & \textbf{12.9}\\
\midrule
\gpta$^\spadesuit$ w/ IKE \cite{zheng2023can} & 15.5 & 8.3 & 1.6 & 1.2 & 17.8 & 11.0 \\
\gpta$^\spadesuit$ w/ ICE \cite{cohen2023evaluating} & 16.1 & 8.5 & 2.1 & 1.4 & 18.1 & 11.5\\
\gpta$^\spadesuit$ w/ MeLLo \cite{zhong2023mquake} & 29.9 & 20.0 & 3.7 & 1.6 & 31.2 & 25.1\\
\gpta$^\spadesuit$ w/ \modelname (Ours) & \textbf{47.2} & \textbf{38.0} & \textbf{51.3} & \textbf{48.0} & \textbf{60.1} & \textbf{53.7}\\
\midrule
\gptb$^\spadesuit$ w/ IKE \cite{zheng2023can} & 18.2 & 8.9 & 1.9 & 1.2 & 19.1 & 10.2\\
\gptb$^\spadesuit$ w/ ICE \cite{cohen2023evaluating} & 19.0 & 8.6 & 2.3 & 1.2 & 19.5 & 10.8\\
\gptb$^\spadesuit$ w/ MeLLo \cite{zhong2023mquake} & 32.2 & 25.6 & 4.7 & 2.3 & 33.5 & 27.8\\
\gptb$^\spadesuit$ w/ \modelname (Ours) & \textbf{49.1} & \textbf{38.8} & \textbf{52.0} & \textbf{49.2} & \textbf{60.9} & \textbf{54.8}\\
\bottomrule
		\end{tabular}
	\end{adjustbox}
	\caption{Experimental results (accuracy; \%) on the dataset MQuAKE-3k 
    and MQuAKE-hard.
    ``$k$ edited'' denotes the KE batch size, which means the number of instances providing new knowledge on answering every question.
    Models with $^\spadesuit$ are closed source.
    The best KE result on every LLM is highlighted in \textbf{bold} font. 
    \label{tab:acc}}
    \vspace{0mm}
\end{table*}

\begin{table}[t]
	\centering
 \small
 \setlength{\tabcolsep}{1.2mm}
		\begin{tabular}{@{}lcccc@{}}
			\toprule
			\textbf{LLM as Constraint Verfier}
                & CON
                &
            COH
                &
            REC
                &
            PER
   \\ 
			\midrule
			\midrule
            \llamaa & 82.3 & 77.9 & 86.3 & 78.2\\
	      \gpta & 92.8 & 91.5 & 92.2 & 86.7 \\
			\bottomrule
		\end{tabular}
	\caption{Experimental results (verification accuracy; \%) of our verifiers for different constraints.
 \vspace{0mm}
    \label{tab:ana_ver}}
\end{table}



\stitle{Model configuration}
We set the hyper-parameters of the baseline methods as suggested by their papers.
We apply the KE methods to the popular LLMs \gpta, \gptb, and \llamaa \cite{chatgpt,touvron2023llama} for evaluation.
The former two are black-box LLMs and the latter one is a white-box LLM.
By default, we set the decoding temperature as 0.0 to minimize the randomness of LLMs' outputs.
Our constraint verifiers use the same LLM as the generator.

\subsection{Overall Performance}\label{sec:4-1}
We incorporate \modelname with several choices of LLMs to test its effectiveness on editing LLMs' knowledge.
We report the test accuracy on \dataa, \datab, \datac in \Cref{tab:acc}.
When the KE batch size is 100, our \modelname improves the accuracy of question answering with KE by 69\%, 58\%, and 52\% over MeLLo on the \gpta, \gptb, \llamaa respectively over the strong baseline method MeLLo on \dataa.
On \datad, our \modelname outperforms the strong baseline method MeLLo by $\times$7, $\times$12, and  $\times$10 times in terms of question answering accuracy with new knowledge on the LLMs \llamaa, \gptb, and \gpta respectively.

When the KE batch is full batch, our \modelname still outperforms the baseline methods and improves the KE accuracy by a significant margin.
%
Our DeepEdit enhances the LLMs' reasoning on new knowledge through higher conciseness, coherence, pertinence, and receptiveness. 
The above advantages are especially significant when the LLMs have to utilize multiple new knowledge to get the correct answers because more new knowledge requires LLMs to be correctly aware of the new knowledge on more reasoning steps. 
The correct knowledge incorporation with our decoding constraints significantly improves the KE performance given more new knowledge.
Overall, our \modelname achieves substantial improvements on KE for LLMs and consistently outperforms the baseline KE methods on various benchmarks.

\begin{table}[t]
	\centering
	\begin{adjustbox}{width=0.99\linewidth}
		\begin{tabular}{@{}lccc@{}}
			\toprule
			\textbf{\gpta w/ \modelname}
                & \textbf{49.1}
                & $\Delta$
   \\ 
			\midrule
			\midrule
			w/o \cona & 46.1  & $\downarrow 6\%$\\
			w/o \conb & 42.9 & $\downarrow 13\%$\\
                w/o \conc& 45.2 & $\downarrow 8\%$\\
                w/o \cond & 46.6 & $\downarrow 5\%$\\
			\bottomrule
		\end{tabular}
	\end{adjustbox}
	\caption{Experimental results (accuracy; \%) on the dataset \dataa of \modelname with KE batch size of 100.
    \vspace{-8mm}
    \label{tab:abl_con}}
\end{table}

\subsection{Accuracy of Constraint Verification}
We conduct an in-depth analysis on the constraint verification of our constraints.
We randomly sample 4,000 pairs of sentences from \dataa that satisfy our constraints as positive samples and the same number of sentence pairs from different instances as the negative samples.
We evaluate different verifiers based on \llamaa and \gpta on the above samples. 
We report the verification accuracy in \Cref{tab:ana_ver}.
We observe high verification accuracy of around 80\% and over 90\% on different constraints achieved by \llamaa and \gpta respectively.
This result matches the performance gap between two LLMs on general reasoning tasks \cite{chatgpt,touvron2023llama}.

\subsection{Efficiency of Depth-first Search} \label{sec:eff_exp}
As analyzed in \Cref{sec:3_5}, we utilize the depth-first search to efficiently increase the LLMs' reasoning depth with new knowledge.
We evaluate the efficiency of our depth-first search using the 3-hop and 4-hop questions in the \datab benchmark, which holds the longest reasoning chains among the instances.

\begin{figure}[!tb]
	\centering
	\includegraphics[width=1.0\linewidth]{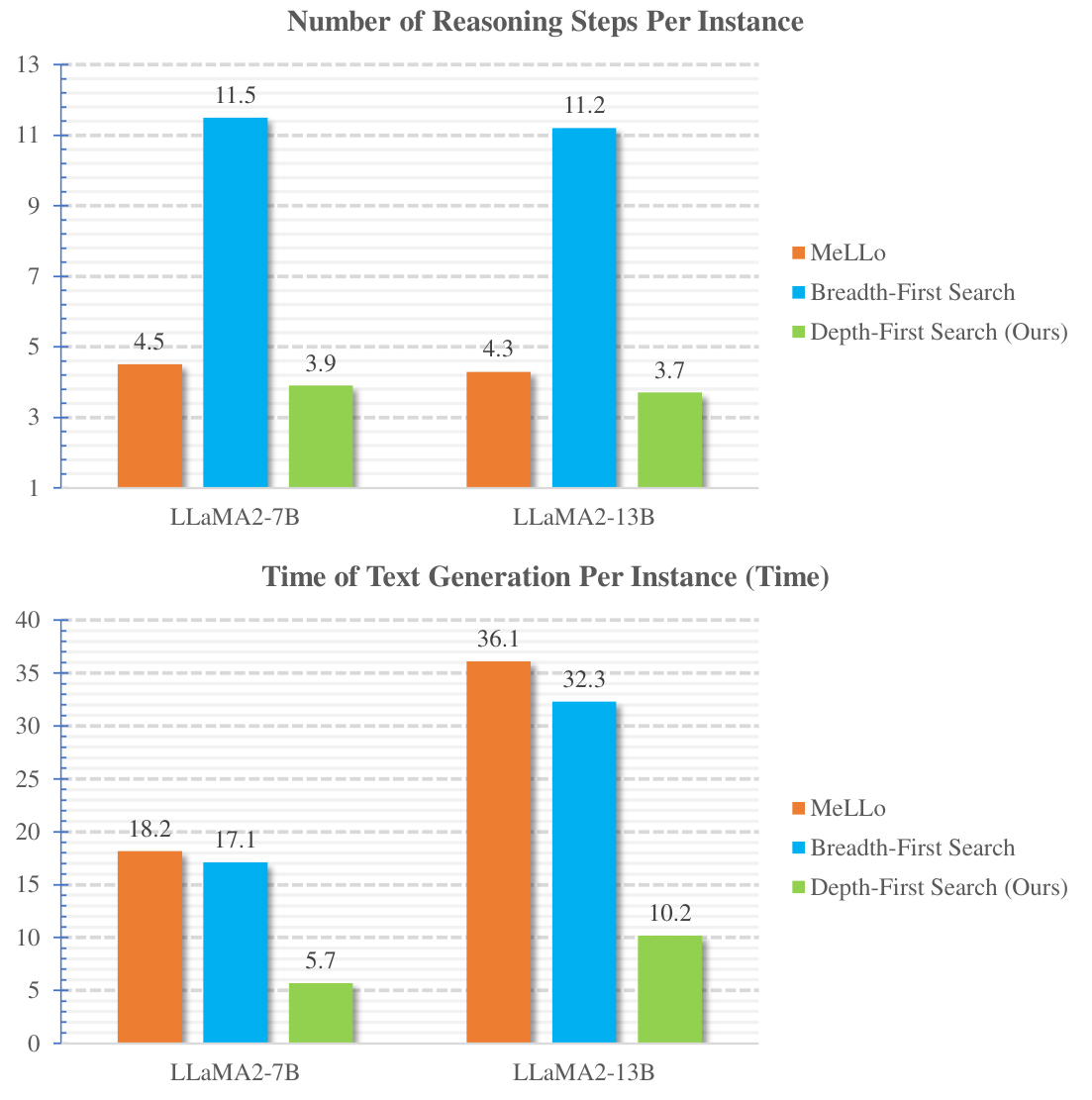}
	\caption{
	Inference costs of the baseline method MeLLo \cite{zhong2023mquake}, breadth-first search introduced in Sec. \ref{sec:3_5}, and our depth-first search.
		\label{fig:cost}}
\end{figure}

The methods we evaluate include: MeLLo, the breadth-first search-based decoding method introduced in Sec. \ref{sec:3_5}, and our depth-first earch.
The average number of reasoning steps and the time of text generation on every instance taken by different methods are reported in \Cref{fig:cost}. where the generation time is on a Linux Server using three A6000 GPUs.

We notice that, compared with the breadth-first search method, our \modelname significantly reduces the number of reasoning steps and the generation time.
Our \modelname efficiently finds the reasoning chain with new knowledge by considering the importance of different step candidates.
These results agree with the theoretical analysis in Sec. \ref{sec:3_5}, since our \modelname' time complexity is $\mathcal{O}(h)$ in the base case, while the time complexity of the breadth-first search is $\mathcal{O}(c^h)$, much higher than our \modelname.
Compared with the baseline MeLLo, our \modelname leads to much less generation time because our output text is purely the reasoning steps without the redundant subquestions and edited facts kept by MeLLo.

\subsection{Ablation Study}
We conduct ablation studies on \modelname to empirically examine the contribution of its main technical components.
We evaluate the influence of different constraints on \modelname.
We report the experimental results of the ablation study of constraints in \Cref{tab:abl_con}. 
We observe that removing each constraint 
leads to substantial performance degradation in different degrees.
This indicates that different constraints contribute to more effective KE from the perspectives of progress, relevance, coherence, and awareness as a whole,
and removing any of them will lead to more mistakes in the reasoning of LLMs.
The above results also validate that the LLMs' outputs cannot automatically satisfy the constraints that represent the desired properties without our \modelname.
More experiments and case studies can be found in the appendix. 

\section{Conclusion}
We explore a new paradigm for knowledge editing of black-box LLMs: decoding with constraints.
We have designed a new decoding method \modelname to improve the knowledge editing of LLMs.
\modelname enhances LLM's outputs to soundly and efficiently incorporate new knowledge in the multi-step reasoning.
In addition to \modelname, we provide two new benchmarks: \datac and \datad to provide more precise and challenging evaluations for knowledge editing methods.
Extensive experiments demonstrate the effectiveness of our \modelname.
Future work includes exploring other constraints that can improve knowledge editing for large language models.

\section*{Limitations}
\modelname is an early effort at conducting knowledge editing in the decoding stage.
There still exist spaces for improving \modelname.
For example, can we have more decoding constraints to improve knowledge editing?
And can we develop more advanced decoding strategies to make LLMs follow the constraints?
We hope the future work can answer above questions and further improve knowledge editing to serve more real-world applications.

\section*{Ethical Considerations}
Ethical considerations are of utmost importance in our research endeavors.  
In this paper, we strictly adhere to ethical principles by exclusively utilizing open-source datasets and employing various models that are either open-source or widely recognized in the scientific community.
We are committed to upholding ethical standards throughout the research process, prioritizing transparency, and promoting the responsible use of technology for the betterment of society.


\bibliography{acl2020}
\bibliographystyle{acl_natbib}

\newpage
\appendix
\section{Analysis on Knowledge Editing Batch Size}

\begin{figure}[!b]
	\centering
	\includegraphics[width=1\linewidth]{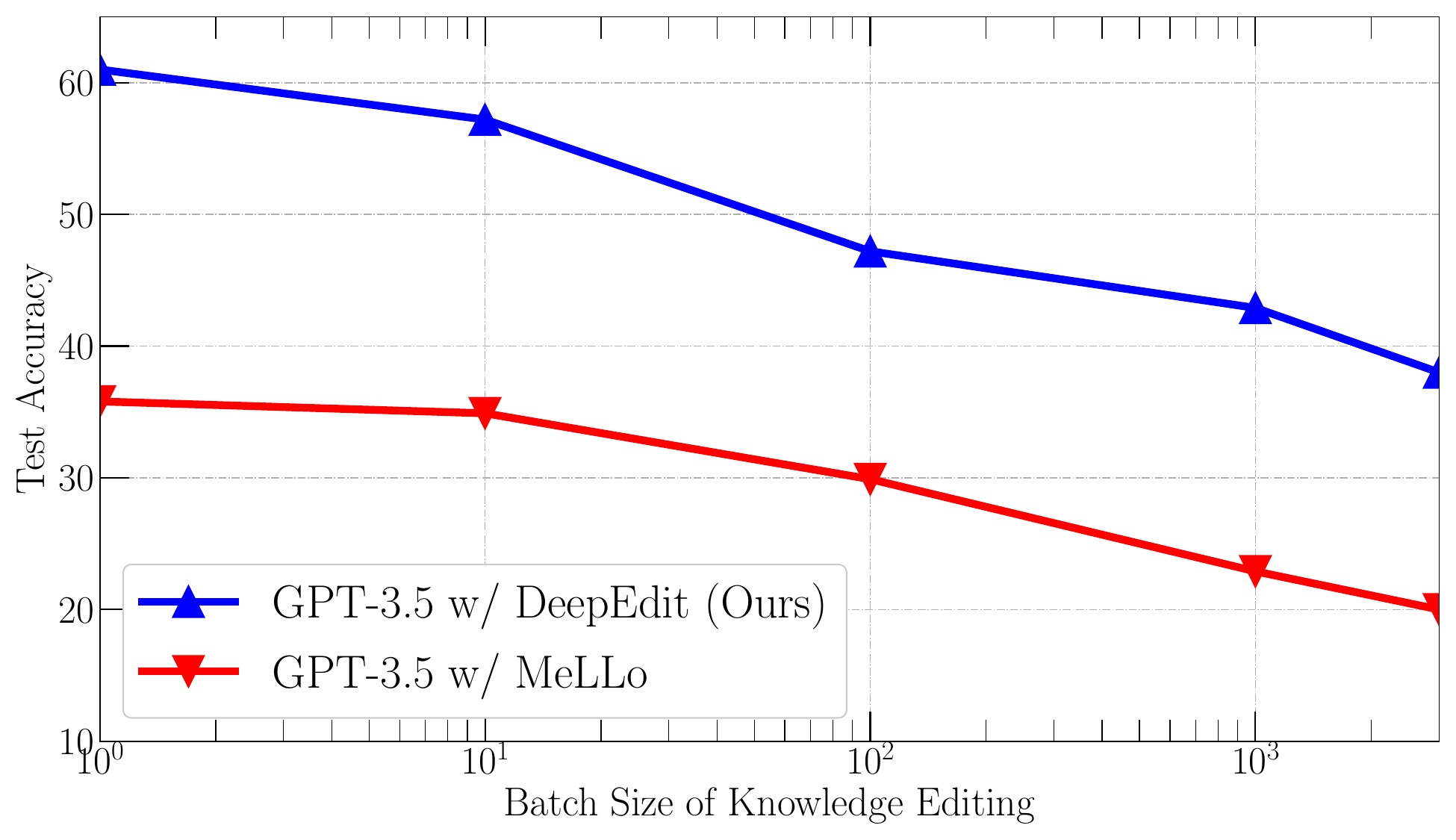}
	\caption{
    Test accuracy (\%) of question answering with new knowledge with different KE batch sizes.
    We present the performance of MeLLo and our \modelname applied on \gpta.
    \vspace{0mm}
\label{fig:result_appendix_1}}
\end{figure}

We follow the existing work \cite{zhong2023mquake} to evaluate the influence of the KE batch sizes on the performance.
A larger batch size leads to the more new knowledge collected from more instances for retrieval. 
We present the test accuracy on \datab with different batch sizes of KE in \Cref{fig:result_appendix_1}.
We observe that the performance of \gpta with MeLLo and our \modelname decreases with a larger batch size of knowledge editing. 
The reason is that with a larger batch, the difficulty of retrieving the relevant new knowledge to the target questions becomes higher.
On the other hand, our \modelname still exhibits significant and consistent improvements over the strong KE method MeLLo.

\begin{figure*}[!tb]
	\centering
	\includegraphics[width=1\linewidth]{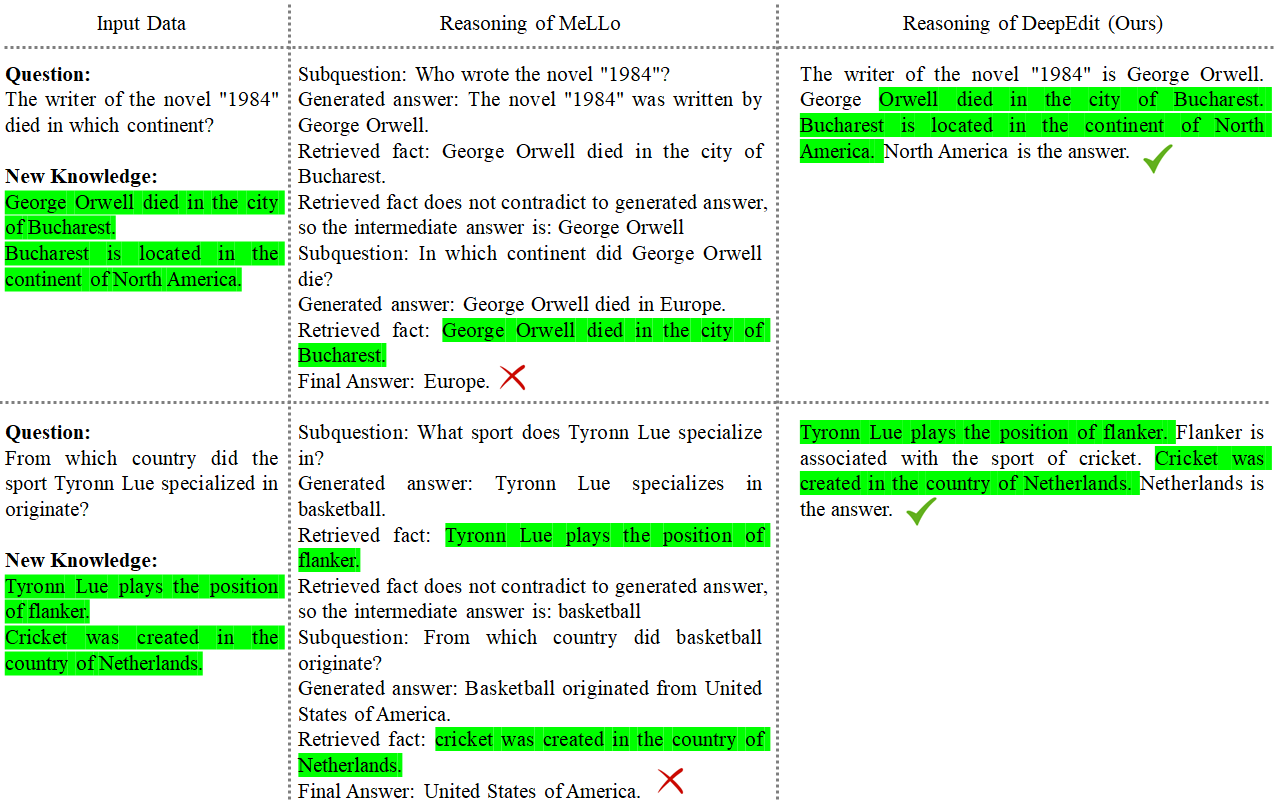}
	\caption{
A case study for MeLLo and our \modelname on the dataset MQuAKE-3k.
The new knowledge are in green.
We report the prediction of \gptb with MeLLo and our \modelname.\label{fig:case}}
\vspace{0mm}
\end{figure*}

\section{Analysis on the Data where the Edited Facts do not Change the Ground-Truth Answers}
Knowledge editing methods enhance the LLMs' reasoning capability on the new knowledge. 
However, there exists the risk that the KE methods can degrade the LLMs' performance on the data where no new knowledge changes their ground-truth answers.
To evaluate whether our method \modelname degrades the performance of LLMs in this case. 
We do the following experiments, for every question-answering instance in \datab, we retrieve the edited facts only from other instances as the new knowledge, which would not change the ground-truth answer for the target question.
In this setting, we first test the \gpta without any KE methods, and compare the performance of \gpta with the KE methods MeLLo and our \modelname.
We present the performance in \Cref{tab:no_edit_1}.

\begin{table}[t]
	\centering
	\begin{adjustbox}{width=\linewidth}
        \setlength{\tabcolsep}{3pt}
		\begin{tabular}{@{}lcccc@{}}
			\toprule
			\textbf{Method}
			& \datab \\ 
			\midrule
			\midrule
			\gpta & 55.7 \\
            \midrule
			\gpta w/ MeLLo \cite{zhong2023mquake} & 54.2 \\
			\gpta w/ \modelname (Ours) & \textbf{55.5} \\
			\bottomrule
		\end{tabular}
	\end{adjustbox}
	\caption{Experimental results (accuracy; \%) on the dataset \datab with no edited facts change the ground-truth answers.
    \label{tab:no_edit_1}}
\end{table}

The empirical results show that our \modelname exhibits higher robustness to give the correct answers for the multi-hop question answering than MeLLo.
The performance drop of \gpta with our \modelname when no edited facts change the ground-truth labels is negligible.

In addition to the above experiments, we also follow existing work \cite{meng2022locating,zheng2023can} to evaluate the Specificity scores of different knowledge editing methods on \datae given the KE batch size as full batch, which measures the accuracy of querying neighborhood prompts by Neighborhood Scores.
In \datae, the neighborhood prompts share the same original object with the target prompt and these facts are not supposed to be edited.
The results are presented in \Cref{tab:no_edit}.

\begin{table}[t]
	\centering
	\begin{adjustbox}{width=\linewidth}
        \setlength{\tabcolsep}{3pt}
		\begin{tabular}{@{}lcccc@{}}
			\toprule
			\textbf{Method}
			& \datae \\ 
			\midrule
			\midrule
			\llamaa & 85.3 \\
            \midrule
            \llamaa w/ MEND \cite{mitchell2021fast}
            & 57.8\\
            \llamaa w/ MEMIT \cite{meng2022mass} & 80.9
            \\
		\llamaa w/ MeLLo \cite{zhong2023mquake} & 82.7 \\
		\llamaa w/ \modelname (Ours) & \textbf{84.1} \\
			\bottomrule
		\end{tabular}
	\end{adjustbox}
	\caption{Specificity Score on the dataset \datae.
    \label{tab:no_edit}}
\end{table}
Our \modelname outperforms all the baseline KE methods in terms of the Specificity Score, which validates that our \modelname causes less performance degradation for LLMs than the baseline methods when the edited facts do not change the ground-truth labels.

\section{Case Study}
\Cref{fig:case} gives a qualitative comparison example between MeLLo and our \modelname on MQuaKE-3k.
The results show that the strong KE method MeLLo makes mistakes on KE.
For example, given the question ``\textit{The writer of the novel ``1984'' died in which continent?
}'', MeLLo is unable to generate the helpful sub-question to find the final correct answer.
In contrast, our \modelname, effectively controls LLMs to correctly find the relevant new knowledge ``Bucharest is located in the continent of North America.'' and obtain the correct answer North America with respect to the new knowledge. 

Taking a closer look, we observe that our \modelname's induced reasoning is much more succinct than MeLLo, since we effectively filter out the irrelevant and redundant information during decoding.
Because longer and noisy textual information can lead to more serious hallucinations and higher difficulty of reasoning, our \modelname exhibits a more effective KE than MeLLo with its succinct reasoning process controlled by constraints.
Overall, our \modelname enhances LLMs to follow our decoding constraints so as to soundly incorporate new knowledge in multi-step reasoning, which leads to the correct predictions in \Cref{fig:case}.

\section{Discussion with MeLLo}
MeLLo \cite{zhong2023mquake} utilizes a prompt engineering based greedy search to produce a reasoning chain that is receptive to new knowledge.
At every iteration, MeLLo retrieves a new knowledge item that is semantically close to the generated reasoning step.
It assesses whether the retrieved new knowledge is conflicted with the generated step.
If so, it takes the new knowledge as next reasoning step, or the generated step otherwise.
Below, we compare the differences between MeLLo and our \modelname from different perspectives:

\begin{itemize}
    \item \textit{Generation Prompt.} As shown in Fig. \Cref{fig: system_prompt_1} and \Cref{fig: system_prompt_2}, MeLLo's generation prompt is much more complex than our \modelname.
    MeLLo leads to much longer context than our \modelname for every inference, which induces more serious hallucinations and time/computational costs.

    \item \textit{Greedy Search versus Depth-First Search}.
    MeLLo employs the greedy search to find the reasoning chain consisting of both parametric and new knowledge.
    If there is error on any reasoning step, the final answer cannot be gotten without the backtracking mechanism.
    We employ the depth-first search to look for satisfactory reasoning chains with new knowledge, of which the backtracking mechanism provides a fault-tolerance mechanism to complement the mistakes happening in any reasoning steps.

    \item \textit{New Knowledge Selection} How and where to use the new knowledge in the reasoning chain is the key to getting the correct answer with new knowledge.
    MeLLo only uses a conflict assessment defined in the prompt template to determine whether to use a new knowledge item.
    In contrast, our \modelname comprehensively evaluates different new knowledge using explicit constraints and a semantic-based importance estimator.
\end{itemize}

\begin{figure*}[htbp]
    \begin{userquery}
    \textbf{Text Generation Prompt of \modelname}: 
    
Question: What is the capital of the country where Plainfield Town Hall is located?
Thoughts with New Knowledge: Plainfield Town Hall is located in the country of the United States of America.\# The capital of United States is Washington, D.C.\# Washington, D.C. is the answer.
Answer: Washington, D.C.
\\

Question: In which country is the company that created Nissan 200SX located?
Thoughts with New Knowledge: Nissan 200SX was created by Nissan.\# Nissan is located in the country of Japan.\# Japan is the answer.
Answer: Japan
\\

Question: Which continent is the country where the director of "My House Husband: Ikaw Na!" was educated located in?
Thoughts with New Knowledge: The director of "My House Husband: Ikaw Na!" is Jose Javier Reyes.\# Jose Javier Reyes was educated at De La Salle University.\# De La Salle University is located in the country of Philippines.\# Philippines is located in the continent if Asia.\# Asia is the answer.
Answer: Asia
\end{userquery}
\caption{Text Generation Prompt of \modelname (Ours).}
\label{fig: system_prompt_1}
\end{figure*}

\begin{figure*}[htbp]
    \begin{userquery}
    \textbf{Text Generation Prompt of MeLLo}: 

Question: What is the capital city of the country of citizenship of Ivanka Trump's spouse?
Subquestion: Who is Ivanka Trump's spouse?
Generated answer: Ivanka Trump's spouse is Jared Kushner.
Retrieved fact: Ivanka Biden is married to Samantha Cameron.
Retrieved fact does not contradict to generated answer, so the intermediate answer is: Jared Kushner
Subquestion: What is the country of citizenship of Jared Kushner?
Generated answer: The country of citizenship of Jared Kushner is United States.
Retrieved fact: Jared Kushner is a citizen of Canada.
Retrieved fact contradicts to generated answer, so the intermediate answer is: Canada
Subquestion: What is the capital city of Canada?
Generated answer:  The capital city of Canada is Ottawa.
Retrieved fact: The capital city of United States of America is Seattle.
Retrieved fact does not contradict to generated answer, so the intermediate answer is: Ottawa
Final answer: Ottawa
\\

Question: Who is the head of state of the country where Rainn Wilson holds a citizenship?
Subquestion: What is the country of citizenship of Rainn Wilson?
Generated answer: Rainn Wilson is a citizen of United States of America.
Retrieved fact: Rainn Wilson is a citizen of Croatia.
Retrieved fact contradicts to generated answer, so the intermediate answer is: Croatia
Subquestion: What is the name of the current head of state in Croatia?
Generated answer: The head of state of Croatia is President Zoran Milanović.
Retrieved fact: The name of the current head of state in Croatia is Kolinda Grabar-Kitarović.
Retrieved fact contradicts to generated answer, so the intermediate answer is: Kolinda Grabar-Kitarović
Final answer: Kolinda Grabar-Kitarović
\\

Question: Who is the spouse of the head of state in United States of America?
Subquestion: Who is the head of state in United States of America?
Generated answer: The US president is Donald Trump.
Retrieved fact: The head of state in United States of America is Joe Biden.
Retrieved fact contradicts to generated answer, so the intermediate answer is: Joe Biden
Subquestion: Who is the spouse of Joe Biden?
Generated answer: The spouse of Joe Biden is Jill Biden.
Retrieved fact: The spouse of Joe Bill is Evan Austin.
Retrieved fact does not contradict to generated answer, so the intermediate answer is: Jill Biden
Final answer: Jill Biden
\\

Question: On which continent is the country of citizenship of the founder of the manufacturer of iPhone 5 situated?
Subquestion: Which company is iPhone 5 produced by?
Generated answer: iPhone 5 is produced by Apple.
Retrieved fact: The company that produced iPhone 5 is Iveco.
Retrieved fact contradicts to generated answer, so the intermediate answer is: Iveco
Subquestion: Who is the founder of Iveco?
Generated answer: Iveco was founded by Giovanni Agnelli.
Retrieved fact: House of Bonaparte was founded by Gustav I of Sweden.
Retrieved fact does not contradict to generated answer, so the intermediate answer is: Giovanni Agnelli
Subquestion: What is the country of citizenship of Giovanni Agnelli?
Generated answer: Giovanni Agnelli is a citizen of Italy.
Retrieved fact: Giovanni Agnelli is a citizen of Niger.
Retrieved fact contradicts to generated answer, so the intermediate answer is: Niger.
Subquestion: On which continent is Niger situated?
Generated answer: Niger is situated on Africa.
Retrieved fact: Kingdom of England is located in the continent of North America.
Retrieved fact does not contradict to generated answer, so the intermediate answer is: Africa.
Final answer: Africa
\end{userquery}
\caption{Text Generation Prompt of MeLLo \cite{zhong2023mquake}.}
\label{fig: system_prompt_2}
\end{figure*}
\end{document}